\documentclass[enabledeprecatedfontcommands]{scrartcl}

\usepackage[ngerman,english]{babel}
\usepackage[T1]{fontenc}
\usepackage{microtype}
\usepackage{glossaries}
\usepackage[detect-all,binary-units]{siunitx}

\usepackage[frozencache,cachedir=.]{minted}
\setminted{fontsize=\small,bgcolor=gray!10}
\setmintedinline{bgcolor=,fontsize=\small}

\usepackage{graphicx}
\usepackage{cleveref}
\usepackage{graphicx}
\usepackage{tikz}
\usetikzlibrary{arrows,arrows.meta,bending,positioning,calc,patterns,backgrounds,fit,decorations.pathreplacing,shapes.geometric}

\usepackage{nicefrac}
\usepackage{booktabs}
\usepackage{listings}
\usepackage{transparent}
\usepackage{multicol}

\definecolor{color0}{HTML}{1f77b4}  
\definecolor{color2}{HTML}{2ca02c}  
\definecolor{color3}{HTML}{d62728}  
\definecolor{color6}{HTML}{e377c2}  
\definecolor{color10}{RGB}{148,103,188}  
\definecolor{color11}{RGB}{255,127,13}  

\usepackage[backend=biber,maxbibnames=1,minbibnames=1,mincitenames=1,firstinits=true,sorting=none,style=ieee,doi=true,eprint=false,isbn=false,url=false,natbib=true]{biblatex}
\DeclareSIUnit{\arbitraryunit}{a.u.}

\newacronym{adc}{ADC}{analog-to-digital converter}
\newacronym{adex}{AdEx}{adaptive exponential integrate-and-fire}
\newacronym{afib}{AF}{atrial fibrillation}
\newacronym{ann}{ANN}{artificial neural network}
\newacronym{asic}{ASIC}{application-specific integrated circuit}
\newacronym{asicab}{\acrshort{asic} adapter \acrshort{pcb}}{\acrlong{asic} adapter \acrlong{pcb}}
\newacronym{api}{API}{application programming interface}
\newacronym{bmbf}{BMBF}{German Federal Ministry of Education and Research}
\newacronym{bptt}{BPTT}{backpropagation through time}
\newacronym{bss2}{\mbox{BSS-2}}{Brain\mbox{ScaleS-2}}
\newacronym{bss1}{\mbox{BSS-1}}{Brain\mbox{ScaleS-1}}
\newacronym{bss2os}{\gls{bss2} OS}{\gls{bss2} Operating System}
\newacronym{bss}{BSS}{BrainScaleS}
\newacronym{cdnn}{CDNN}{convolutional deep neural network}
\newacronym{cpu}{CPU}{central processing unit}
\newacronym{dfki}{DFKI}{German Research Centre for Artificial Intelligence}
\newacronym{dma}{DMA}{direct memory access}
\newacronym{dram}{DRAM}{dynamic random-access memory}
\newacronym{ecg}{ECG}{electrocardiogram}
\newacronym{fpga}{FPGA}{field-programmable gate array}
\newacronym{gbe}{GbE}{gigabit ethernet}
\newacronym{i2c}{I\textsuperscript{2}C}{Inter-Integrated Circuit}
\newacronym{ic}{IC}{integrated circuit}
\newacronym{isa}{ISA}{instruction set architecture}
\newacronym{itl}{ITL}{in-the-loop}
\newacronym{jit}{JIT}{just-in-time}
\newacronym{lvds}{LVDS}{low-voltage differential signaling}
\newacronym{lif}{LIF}{leaky-integrate and fire}
\newacronym{li}{LI}{leaky integrator}
\newacronym{mac}{MAC}{multiply–accumulate}
\newacronym{madc}{MADC}{membrane \acrshort{adc}}
\newacronym{mse}{MSE}{mean squared error}
\newacronym{cadc}{CADC}{columnar \acrshort{adc}}
\newacronym{pcb}{PCB}{printed circuit board}
\newacronym{ppu}{\acrshort{simd} \acrshort{cpu}}{\acrlong{simd} \acrlong{cpu}}
\newacronym{relu}{ReLU}{rectified linear unit}
\newacronym{rtl}{RTL}{Register Transfer Level}
\newacronym{gd}{GD}{gradient descent}
\newacronym{simd}{SIMD}{single instruction, multiple data}
\newacronym{snn}{SNN}{spiking neural network}
\newacronym{sodimm}{\mbox{SO-DIMM}}{small outline dual in-line memory module}
\newacronym{sram}{SRAM}{static random-access memory}
\newacronym{stdp}{STDP}{spike timing dependent plasticity}
\newacronym{stp}{STP}{short term plasticity}
\newacronym{rnn}{RNN}{recurrent neural network}
\newacronym{rsnn}{RSNN}{recurrent spiking neural network}
\newacronym{nasprop}{NASProp}{neuromorphic accumulative spike propagation}
\newacronym{vu}{VU}{vector unit}
\newacronym{udp}{UDP}{user datagram protocol}
\newacronym{cd}{CD}{continuous deployment}
\newacronym{ci}{CI}{continuous integration}
\newacronym{hpc}{HPC}{high-performance computing}
\newacronym{gpu}{GPU}{graphics processing unit}
\newacronym{usb}{USB}{universal serial bus}
\newacronym{sfnn}{SFNN}{spiking feed-forward neural network}
\newacronym{sfnnwlrf}{SFNNwLRF}{\gls{sfnn} with limited receptive field}
\newacronym{ttfs}{TTFS}{time-to-first spike}
\newacronym{scnn}{SCNN}{spiking convolutional neural network}
\newacronym{srnn}{SRNN}{spiking recurrent neural network}
\newacronym{dsnn}{DSNN}{deep spiking neural network}
\newacronym{dnn}{DNN}{deep neural network}
\newacronym{prng}{PRNG}{pseudorandom number generator}

\begin{document}

\title{Scalable Network Emulation on\\Analog Neuromorphic Hardware}

\author{%
	Elias~Arnold\thanks{European Institute for Neuromorphic Computing, Kirchhoff-Institute for Physics, Heidelberg University, Germany}
	\and
	Philipp~Spilger\footnotemark[1]
	\and
	Jan~V.~Straub\footnotemark[1]
	\and
	Eric~Müller\footnotemark[1]\\
	\and
	Dominik~Dold\thanks{Advanced Concepts Team, European Space Research and Technology Centre, European Space Agency, The Netherlands}
	\and
	Gabriele~Meoni\thanks{Faculty of Aerospace Engineering, Delft University of Technology, The Netherlands}
	\and
	Johannes~Schemmel\footnotemark[1]
}

\date{2024-11-05}

\maketitle

\begin{abstract}
	We present a novel software feature for the \acrlong{bss2} accelerated neuromorphic platform that facilitates the partitioned emulation of large-scale \acrlongpl{snn}.
	This approach is well suited for \acrlongpl{dsnn} and allows for sequential model emulation on undersized neuromorphic resources if the largest recurrent subnetwork and the required neuron fan-in fit on the substrate. 
	The ability to emulate and train networks larger than the substrate provides a pathway for accurate performance evaluation in planned or scaled systems, ultimately advancing the development and understanding of large-scale models and neuromorphic computing architectures.
	We demonstrate the training of two \acrlong{dsnn} models ---using the MNIST and EuroSAT datasets--- that exceed the physical size constraints of a single-chip \acrlong{bss2} system.
\end{abstract}

\glsunset{cpu}
\glsunset{gpu}
\glsunset{usb}
\glsunset{api}

\section{Introduction}\label{sec:introduction}

For traditional deep learning algorithms, whether simulated on conventional hardware or accelerated using \glspl{gpu} and specialized hardware, the seamless integration of machine learning frameworks such as PyTorch and TensorFlow has simplified modeling and accelerated research.
Recent years have seen a parallel evolution in the field of \glspl{snn}, where specialized modeling interfaces~\citep{pehle2021norse,manna2023frameworks} have begun to play a key role in streamlining the model development process.
While the creation of a scaffold for building software support within machine learning libraries for general-purpose processing units is well established~\citep{torchxla,lattner2021mlir}, it is still an open research topic in the context of custom digital neuromorphic hardware~\citep{shrestha2022survey}, and even more so for the time-continuous nature of many analog neuromorphic systems, where the path to seamless integration is considerably more intricate.

In this work, we address typical model size limitations imposed by small substrates such as the \gls{bss2} accelerated mixed-signal neuromorphic system~\citep{pehle2022brainscales2_nopreprint_nourl}, which is currently only deployed in its single-chip variant.
Initially, the \gls{bss2} architecture has been designed as a research vehicle for computational neuroscience, offering specialized features tailored to address the intricacies of neural dynamics and plasticity.
The inclusion of multi-compartmental neurons, complex synapse dynamics, \gls{adex} compartment dynamics~\citep{brette2005adaptive,billaudelle2022accurate}, as well as short-term and long-term plasticity, positions \gls{bss2} as a versatile platform for exploring diverse neural phenomena.
Beyond computational neuroscience, \gls{bss2} also extends its reach into machine-learning-inspired applications, where functional modeling often draws inspiration from machine learning.

\Glspl{dnn} are often significantly larger than neuromorphic \acrshortpl{asic}.
While small-scale multi-chip system prototypes using an EXTOLL-based FPGA-mediated interconnect have been demonstrated~\citep{thommes2022demonstrating,thommes2023phdthesis}, production \gls{bss2} system resources operate in single-chip configurations.
However, networks with limited fan-in requirements that either comprise a pure feed-forward topology or sufficiently local recurrence allow for the partitioning into subnetworks that individually fit onto single \acrshortpl{asic}.
In general, partitioning introduces sequence points where emulation can be paused while the data flow still determines the execution order, i.e.\ subnetwork partitions of early layers are emulated before later layers, but the execution order within a layer is arbitrary.
This therefore enables the sequential evaluation of networks larger than the existing neuromorphic substrate without having to resort to software simulation.
Especially with regard to the typical costs and time required for hardware development, this enables early analysis and thus optimization of future hardware substrates.
The reuse of ``computational units'' (neurons, synapses, routing, and other resources) is analogous to the way conventional von-Neumann architectures utilize computational resources and can be understood as a form of virtualization of the neuromorphic substrate.
This departs from traditional neuromorphic systems, which allocate dedicated resources for each component of spiking neural networks.
Recent work~\citep{mysore2022hierarchical} laid out a partitioning method for mapping large-scale neural network models onto neuromorphic hardware.
Along these lines, for hardware supporting non-time-continuous operation, \citet{song2020compiling} describes a complete workflow from model specification to hardware execution.
Previous work by the authors provided similar functionality for the activation-based ---i.e.\ non-spiking--- operation mode of \gls{bss2}~\citep{spilger2020hxtorch}.

The \gls{bss2} software stack aims to provide a user-friendly modeling \gls{api} that abstracts away from hardware-specific intricacies~\citep{mueller2022scalable_noeprint}.
Over the course of its development, machine learning inspired training approaches have become increasingly popular.
However, until recently, our modeling efforts were mostly limited to the size constraints of single \gls{bss2} \acrshortpl{asic}.
In this work, we focus on providing a framework for integrating such partitioning methods more generally, particularly for large-scale \glspl{snn}, into the \gls{bss2} software stack.
The method not only applies to single-chip substrates, but generalizes also to larger substrates by concurrently placing multiple partitions.

In the following, we focus on scenarios, such as feed-forward networks or those with sufficiently small recurrent subnetworks, where hardware reuse becomes a practical proposition.
The overarching goal is to automate the process of making \gls{bss2} amenable to such cases, thereby extending its capabilities to emulate larger-than-substrate-sized networks efficiently and seamlessly.
Finally, we demonstrate the training and emulation of larger, multi-partition networks on single-chip \gls{bss2} substrates using the MNIST~\citep{lecunmnist} dataset of handwritten digits and the EuroSAT~\citep{helber2017eurosat} dataset for land use and land cover classification.
The latter is of particular relevance for future applications in space~\citep{izzo2022neuromorphic}, as energy-efficient compute infrastructure such as neuromorphic hardware represents a promising candidate for neural solutions onboard spacecraft --- especially miniaturized ones like CubeSats.
We present the first results on \gls{bss2} for training functional networks larger than the hardware substrate.

\section{Methods}\label{sec:methods}

\vspace{-0.0cm}  
\begin{figure}
    \centering
    \tikzset{
             panel/.style={
                     inner sep=0pt, outer sep=0, execute at begin node={\tikzset{anchor=center, inner sep=.33333em}}
             },
             label/.style={anchor=north west, inner sep=0, outer sep=0}
        }
        \begin{tikzpicture}
                \node[panel, anchor=north west
                ] (a) at (0,  3.9) {
                    \includegraphics[width=.32\textwidth]{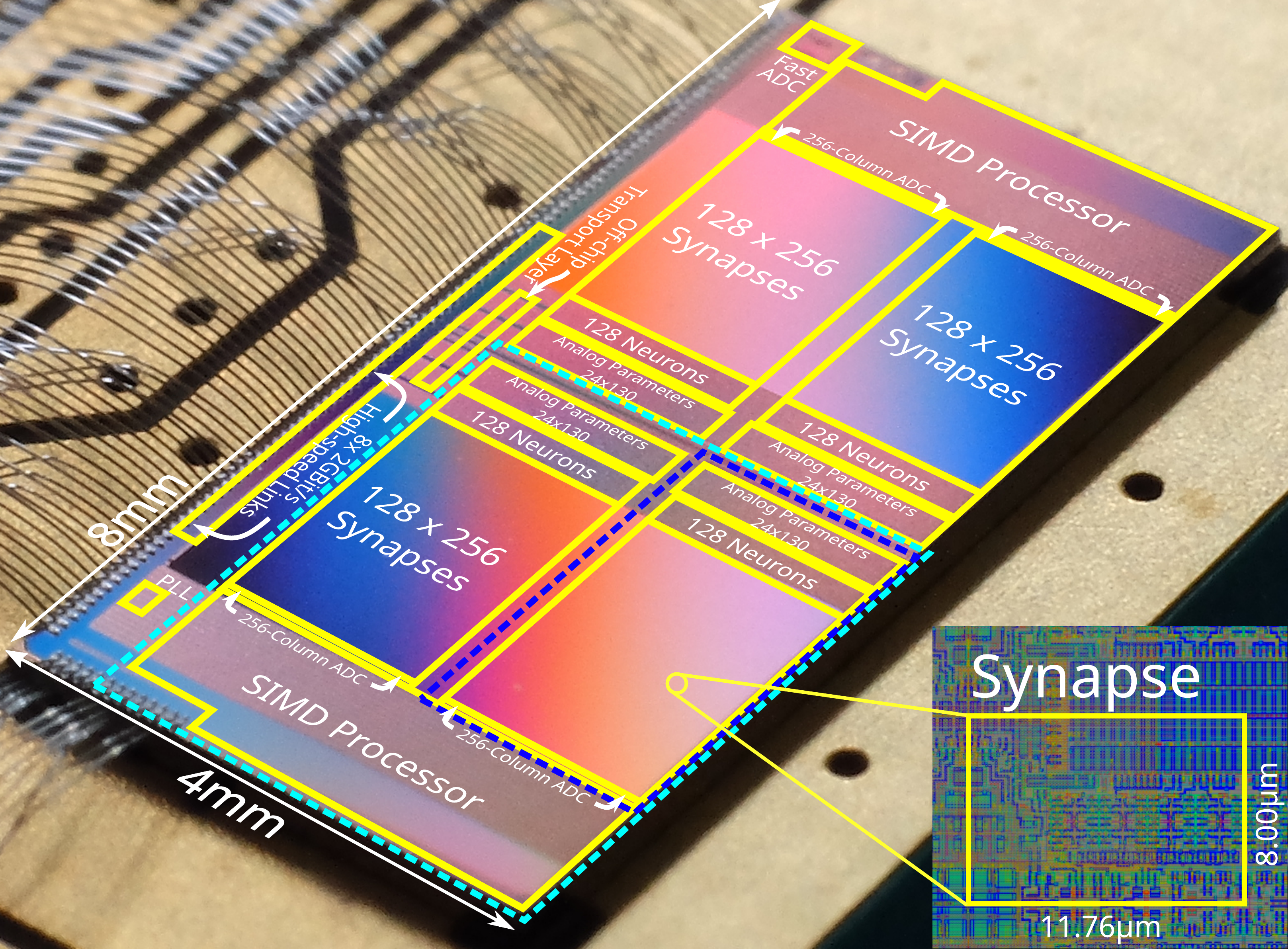}
                };
                \node[label] at (0.2, 3.8) {\textcolor{black}{A}};
                \node[panel, anchor=north west
                ] (b) at (4.9,  3.5) {
\begingroup%
  \makeatletter%
  \providecommand\color[2][]{%
    \errmessage{(Inkscape) Color is used for the text in Inkscape, but the package 'color.sty' is not loaded}%
    \renewcommand\color[2][]{}%
  }%
  \providecommand\transparent[1]{%
    \errmessage{(Inkscape) Transparency is used (non-zero) for the text in Inkscape, but the package 'transparent.sty' is not loaded}%
    \renewcommand\transparent[1]{}%
  }%
  \providecommand\rotatebox[2]{#2}%
  \newcommand*\fsize{\dimexpr\f@size pt\relax}%
  \newcommand*\lineheight[1]{\fontsize{\fsize}{#1\fsize}\selectfont}%
  \ifx\svgwidth\undefined%
    \setlength{\unitlength}{116.34214386bp}%
    \ifx\svgscale\undefined%
      \relax%
    \else%
      \setlength{\unitlength}{\unitlength * \real{\svgscale}}%
    \fi%
  \else%
    \setlength{\unitlength}{\svgwidth}%
  \fi%
  \global\let\svgwidth\undefined%
  \global\let\svgscale\undefined%
  \makeatother%
  \begin{picture}(1,0.88414444)%
    \lineheight{1}%
    \setlength\tabcolsep{0pt}%
    \put(0.1891335,0.07585879){\color[rgb]{0,0,0}\makebox(0,0)[lt]{\lineheight{1.25}\smash{\begin{tabular}[t]{l}\footnotesize Recurrent\\\\\end{tabular}}}}%
    \put(0.19115407,0.01160621){\color[rgb]{0,0,0}\makebox(0,0)[lt]{\lineheight{1.25}\smash{\begin{tabular}[t]{l}\footnotesize Feed forward\\\\\end{tabular}}}}%
    \put(0,0){\includegraphics[width=\unitlength,page=1]{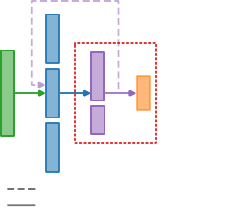}}%
  \end{picture}%
\endgroup%

                };
                \node[label] at (5.0, 3.8) {B};
                \node[panel, anchor=north west
                ] (c) at (7.9,  3.8) {
					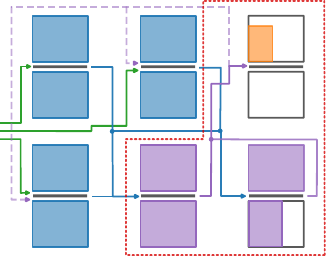
                };
                \node[label] at (7.8,  3.8) {C};
                \node[panel, anchor=north west
                ] (d) at (14,  3.7) {
					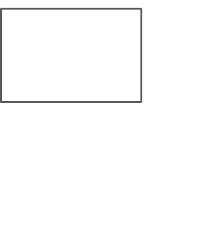
                };
                \node[label] at (13.8,  3.8) {D};
                \node[panel, anchor=north west
                ] (d) at (0,  -0.8) {
                \begin{minipage}{0.98\linewidth}
\begin{lstlisting}[
                            basicstyle=\tt\tiny,
                            escapeinside=\'\',
                            numbers=left,
                            xleftmargin=3em,
                            numberstyle=\footnotesize\color{black!30},
                            multicols=2,
                            stepnumber=1]
'\tt\color{color10}first'  = hxtorch.snn.'\tt\color{color10}ExecutionInstance'()
'\tt\color{color10}second' = hxtorch.snn.'\tt\color{color10}ExecutionInstance'()
'\tt\color{color11}third'  = hxtorch.snn.'\tt\color{color11}ExecutionInstance'()

exp  = hxtorch.snn.Experiment()

'\tt\color{color3}syn1' = hxtorch.snn.'\tt\color{color3}Synapse'(exp, '\tt\color{color10}first', ...)
'\tt\color{color3}lif1' = hxtorch.snn.'\tt\color{color3}LIF'(exp, '\tt\color{color10}first', ...)
'\tt\color{color3}syn2' = hxtorch.snn.'\tt\color{color3}Synapse'(exp, '\tt\color{color10}second', ...)
'\tt\color{color3}lif2' = hxtorch.snn.'\tt\color{color3}LIF'(exp, '\tt\color{color10}second', ...)
'\tt\color{color3}syn3' = hxtorch.snn.'\tt\color{color3}Synapse'(exp, '\tt\color{color11}third', ...)
'\tt\color{color3}syn4' = hxtorch.snn.'\tt\color{color3}Synapse'(exp, '\tt\color{color11}third', ...)
'\tt\color{color3}li3'  = hxtorch.snn.'\tt\color{color3}LI'(exp, '\tt\color{color11}third', ...)
'\tt\color{color2}x1' = '\tt\color{color3}syn1'('\tt\color{color2}input')
'\tt\color{color2}x2' = '\tt\color{color3}syn2'('\tt\color{color2}input')
'\tt\color{color2}x3' = '\tt\color{color3}lif1'('\tt\color{color2}x1')
'\tt\color{color2}x4' = '\tt\color{color3}lif2'('\tt\color{color2}x2')
'\tt\color{color2}x5' = '\tt\color{color3}syn3'('\tt\color{color2}x3')
'\tt\color{color2}x6' = '\tt\color{color3}syn4'('\tt\color{color2}x4')
'\tt\color{color2}x7' = '\tt\color{color3}li3'('\tt\color{color2}x5')
'\tt\color{color2}x7' = '\tt\color{color3}li3'('\tt\color{color2}x6')

hxtorch.snn.'\tt\color{color0}run'(exp, ...)

'\tt\color{color6}loss' = f('\tt\color{color2}x7')
'\tt\color{color6}loss'.backward()
\end{lstlisting}
                \end{minipage}
                };
                \node[label] at (0.1, -0.5) {E};
        \end{tikzpicture}
		\caption{%
			\textbf{(A)} A photo of the \gls{bss2} chip with its schematic overlaid on top.
			\textbf{(B)} A larger-scale network, exceeding the size of a single \gls{bss2} substrate.
			To emulate the full network, it can be partitioned into smaller subnetworks and executed concurrently on a multi-chip setup as displayed in \textbf{(C)} or all subnetworks are emulated sequentially by reusing the same chip resource.
			The concept of sequential execution also applies to networks that exceed scaled multi-chip system in size where the scaled system then becomes the largest sequentially allocatable entity.
			Dashed lines correspond to recurrent dependencies.
			\textbf{(D)} Upper: On \gls{bss2} convolutions need to be unrolled spatially thereby demanding excessive hardware resources and partitioning.
			Here, $W$ and $H$ corresponds to the width and height of the kernel, $C^\text{i}$ and $C^\text{o}$ are the number of input resp.\ output feature planes.
			Lower: For sequential network emulation, recurrent dependencies need to fit on a single substrate which reduces external fan-in.
			However, this limitation does not apply for concurrent network emulation.
			\textbf{(E)} Software \gls{api} of explicitly partitioned network indicated by the dotted red line in (B) and (C).
			\label{fig:concepts}
		}
\end{figure}

\noindent 
In this work, the latest \gls{bss2} \acrshort{asic} \citep{pehle2022brainscales2_nopreprint_nourl} is used as a mixed-signal neuromorphic substrate, depicted in \cref{fig:concepts}A.
It features \num{512} (single-compartment) neurons implementing the \gls{adex} model which can receive events via \num{256} synapses each.
Events are propagated via digital signals, while the post-synaptic neuron dynamics evolve in the analog domain.
Using the current default \acrshort{fpga}-\acrshort{asic} link speed, the maximum sustained bandwidth is \SI{250}{\mega\hertz} for both input and output events.
Therefore, the emulation operates time-continuously and in real-time --- in contrast to digital simulation, the experiment in general cannot be paused.
Hence, the network size which can be concurrently (and interdependently) emulated is limited by the number of neuron and synapse circuits, and other resources.
However, concurrent placement and emulation is only required for tightly-coupled recurrent network subgraphs, while feed-forward network subgraphs can be partitioned and run in parts.
\Cref{fig:concepts}C
sketches the partitioning of the feed-forward network in \cref{fig:concepts}B.
Using a multi-chip substrate, the network can be emulated in continuous time.
If there are no recurrent inter-chip dependencies (omitting dotted line in \cref{fig:concepts}B), the inter-chip communication does not need to happen in real time, and can be buffered.
In that case, the whole network can also be emulated sequentially by reusing a single chip.
Since convolutions need to be spatially unrolled on \gls{bss2} (see \cref{fig:concepts}D), spiking convolutional networks on \gls{bss2} will benefit from the presented feature.

Splitting networks into multiple partitions and emulating them sequentially requires the events in-between executions to be recorded and played back in dependent executions.
This increases the required communication of events from and to the system compared to direct forwarding of events within the hardware.
However, for typical machine-learning-inspired training the readout of events from hidden layers is required in any case.

Partitioning projections does not necessarily decrease the fan-in for the post-synaptic layer, since neuron dynamics are not linear.
Thus, we take advantage of the hardware's ability to combine neuron circuits, resulting in an increased fan-in capability of `$256 \cdot \text{\#neuron circuits per neuron}$', up to $256 \times 64 = 16384$ unsigned weights.
We use two \SI{6}{\bit}-weight hardware synapses to represent a signed weight, therefore the maximum number of signed input weights is \num{8192}.
Consequently, this decreases the number of `logical' neurons available per single execution by $\text{\#neurons} = \nicefrac{512}{\text{\#neuron circuits per neuron}}$, possibly increasing the number of required partitions.

We base our work on the existing \gls{bss2} software stack, which provides multiple abstraction layers, see \citet{mueller2022scalable_noeprint} for details.
Specifically, we integrated partitioned execution functionality into the layer that represents experiments as a signal flow graph.
Even before this support was added, the signal flow graph had an understanding of data input and output operations, so the addition of temporary readout and data reinsertion functionality was a natural extension.
To take advantage of developments in the machine learning community, the user-facing hxtorch \gls{api}~\citep{spilger2023hxtorchsnn} is based on PyTorch data structures and integrates with its auto-differentiation functionality.

\subsection{Training}\label{sec:methods:training}

The MNIST and EuroSAT models are trained using well-established surrogate gradient-based learning methods~\citep{neftci2019surrogate}.
Class scores are optimized by minimizing the cross-entropy loss, using the Adam optimizer~\citep{kingma2014adam} with (surrogate) gradients obtained by the \gls{bptt} algorithm.
To approximate the networks' gradients on \gls{bss2}, we apply the hardware-\gls{itl} training procedure \citep{schmitt2017hwitl_nourl} and record and read out the network observables, i.e.\ membrane voltages and spikes.
These observables are mapped to PyTorch tensor data structures with a fixed time grid with resolution $\delta t$.
For this, we calculate the factor which scales the membrane dynamics on \gls{bss2} to the corresponding dynamics in software, that are idealized for gradient estimation.
Synapse and neuron dynamics are numerically integrated on this time lattice in the case of simulated (sub-)networks.
Each part of the network is run, or simulated respectively, for $T=\SI{30}{\micro\second}$ in the case of MNIST and \SI{64}{\micro\second} for the EuroSAT task per image.
The measured/simulated membrane traces $v_k$ in the readout layer are converted into scores $s_k$ via a max-over-time decoding, $s_k = \max_t(v_k(t))$~\citep{cramer2022surrogate} for MNIST, or by taking the last observed membrane value $s_k = v_k( T)$ for the EuroSAT dataset.
The partitioning of the considered \glspl{snn} is explained in \cref{sec:results}.

\subsection{MNIST\label{sec:mnist_methods}}

The MNIST~\citep{lecunmnist} dataset contains \num{70000} $28 \times 28$ gray scale images of handwritten digits that are categorized into \num{10} classes (0 to 9).
\num{60000} images are meant for training purposes, the rest for testing the model.
We consider a fully connected feed-forward network with \num{256} \gls{lif} units in the hidden layer and \num{10} \glspl{li} in the readout layer.
A \gls{ttfs} encoding scheme, described in \cref{sec:results_mnist},
transfers the images from a pixel-value representation to spike events.
The dataset is augmented by using random rotations up to \SI{25}{\degree} which are applied with a probability of \SI{50}{\percent}, additionally we normalize images.
For improved generalization we also use dropout with a probability of \SI{15}{\percent} in the hidden layer, resulting in some of the hidden spikes not being injected into the readout layer during training.
To keep the network's dynamics and parameters within the system capabilities, we use regularization terms for the firing rate in the hidden layer which might exceed the system's bandwidth, the readout membrane traces which might saturate due to the limited range of the \gls{cadc} and the weights which are also limited in range on hardware.
The training process spans $100$ epochs during which the learning rate and firing rate regularization constant decrease exponentially.
At the end of the training, the model's performance is evaluated with the test set.
The final performance is the averaged over different \gls{prng} seeds.
A summary of the used training and model parameters is given in \cref{tab:mnist_params}.

\subsection{EuroSAT}\label{sec:methods:eurosat}

The EuroSAT dataset consists of \num{27000} $64 \times 64 \times 3$ RGB\footnote{We only consider the RGB bands out of the 13 provided spectral bands.} images of the Earth's surface taken by the satellite mission Sentinel-2, categorized into \num{10} classes.
We split the dataset in training, validation, and test set by ratios \num{0.7}, \num{0.1}, and \num{0.2}.
For regularization, random flips are applied to the training images.
For its classification, we consider a network with two hidden \gls{lif} layers consisting of 484 and 128 units, and one \gls{li} readout layer to infer decisions.
For spike encoding of the input images we use a \gls{ttfs} encoding, described by \cref{eq:encoding_eurosat}.
In addition to the training procedure outlined in \cref{sec:methods:training}, we halve the learning rate after the epochs $\lbrace 10, 20, \dots, 60 \rbrace$.
Training is performed for a maximum of \num{500} epochs in simulations, or \num{100} on \gls{bss2}.
If there is no improvements on the validation accuracy for \num{25} epochs in simulation or \num{15} epochs on \gls{bss2}, the training is stopped.
We save the best performing model on the validation set and use it for later evaluation on the test set.
A summary of all model and training parameters is given in \cref{tab:eurosat_params}.

\section{Results}\label{sec:results}

In this section we describe our implementation, which introduces software support for model partitioning and sequential execution on \gls{bss2}.
We demonstrate its use on the MNIST and EuroSAT datasets.

\subsection{Software}

While the user of a machine learning framework does not need to know the partitioning, this information is required in the intermediate representation used for scheduling execution on the hardware.
In the high-level experiment description, networks are comprised of populations of neurons and projections of synapses.
We use a signal-flow graph to represent multiple executions and their data-flow dependencies.
This representation can be used to represent partitioned networks.
To this end, network entities are annotated with information regarding their associated execution (\texttt{ExecutionInstance} in \cref{fig:concepts}E).
The inter-execution projection represents the forwarding of events from one execution to another.
It receives recorded events from the source execution and injects these events into the target execution.
The host computer is used for the translation of the events, which allows for the complete decoupling of event routing constraints between executions.

In our machine learning frontend \texttt{hxtorch.snn}, each layer is assigned to a specific execution via a parameter upon construction.
The inter-execution dependencies are then automatically extracted from the network topology.
This enables explicit (manual) partitioning as well as employing user-defined partitioning algorithms, which can also be used for mixed hardware-emulated and software-simulated networks, see \cref{subsec:eurosat}.
\Cref{fig:concepts}E shows a frontend \gls{api} example.
The data flow used for MNIST classification is visualized in \cref{fig:mnist}B.

It is anticipated that the utilization of multiple partially sequential executions and the increased required data transfer when using multiple partitions in contrast to executing a network in a single hardware run will result in a reduction in runtime performance.
The hardware runtime scales linearly with the depth of the partitioned network, since these executions are required to be run sequentially due to inter-partition data dependencies.
Partitions without data dependencies, e.g., multiple partitions of the same layer, can be executed concurrently.
The choice of whether to execute the partitions concurrently or sequentially depends on the available hardware resources.
Therefore, runtime additionally scales linearly with the ratio of concurrently executable partitions to available hardware.
When using partitioning, all events between partitions are recorded and translated on the host computer.
In contrast, networks executed in a single non-partitioned hardware run only require complete event recording during training,
as only the data from the last layer is typically of interest during inference.
In addition, event recording and translation overhead is expected to impair runtime performance in comparison to non-partitioned experiments.
A dedicated inter-execution memory buffer in some \gls{fpga}-managed \gls{dram} could at least eliminate the software overhead at the cost of additional \gls{fpga} development effort to support additional translation and playback of recorded data.
\Cref{tab:results_performance} shows wall-clock runtime measurements of the MNIST experiment, cf.~\cref{sec:results_mnist}, broken down to evaluate the performance impairment attributed to partitioned execution.
Here, membrane potential recording dominates the hardware runtime, which is potentiated by the linear scaling with the number of partitions.
Event recording and playback via the host computer on the other hand is insignificant.

\begin{table}
	\centering
	\begin{tabular}{lS[table-format=1.1,table-space-text-post=\si{\ms}]l}
	\toprule
	experiment step & {duration} & data \\
	\midrule
	host computer compilation \& post-processing & 692\,\si{\ms} & \\
	\qquad event encoding & 0.3\,\si{\ms} & 721 spikes\\
	\qquad event decoding & 0.7\,\si{\ms} & 909 spikes\\
	\qquad membrane recording decoding & 100\,\si{\ms} & 8445 samples\\
	hardware experiment total & 248\,\si{\ms} & \\
	ML front end data handling, backward pass & 810\,\si{\ms} &\\
	total & 1800\,\si{\ms} & \\
	\midrule
	partitioned hardware runtime (5 partitions) & 40\,\si{\ms} & \\
	\qquad realtime hardware runtime (per partition) & 3\,\si{\ms} & \\
	\qquad inter-batch-entry hardware wait (per partition) & 5\,\si{\ms} & \\
	\bottomrule
	\end{tabular}
	\caption{\label{tab:results_performance}
		Wall-clock duration measurements (top) and user-requested minimal realtime runtimes (bottom) for the model classifying MNIST, cf.\ \cref{sec:results_mnist}, for a single batched execution of \num{100} images with \SI{30}{\us} experiment runtime each.
		In-between batch entries, for relaxing the analog neuron dynamics, a wait period of \SI{50}{\us} is added additionally, resulting in a minimal hardware runtime of \SI{8}{\ms}.
		Since the model is partitioned into five sequential executions, this minimal runtime is scaled linearly to \SI{40}{\ms}.
		The difference to the measured total hardware runtime of \SI{248}{\ms} is attributed predominantly to recording the neuron's membrane potential during the experiment, which also additionally yields \SI{100}{\ms} of host computer runtime.
		Event decoding is required for training, only event encoding of \SI{0.3}{\ms} is attributed to partitioning and sequential execution, which is deemed insignificant.
		While this results in an overall overhead of a factor of \num{600} (or \num{225} when accounting for the relaxation/wait time) between the minimal experiment runtime and the training wall-clock runtime using partitioning, we expect the same experiment to run a factor of five faster (same as number of partitions) on a sufficiently large substrate that allows training without partitioned sequential execution.
	}
\end{table}

\subsection{Examples}
We exemplify our support for partitioning using \gls{snn} models with topologies that otherwise would not be emulatable on a single-chip \gls{bss2} system.

\subsubsection{MNIST}\label{sec:results_mnist}
\begin{figure}
    \centering
    \tikzset{
             panel/.style={
                     inner sep=0pt, outer sep=0, execute at begin node={\tikzset{anchor=center, inner sep=.33333em}}
             },
                label/.style={anchor=north west, inner sep=0, outer sep=0}
        }
        \begin{tikzpicture}
                \node[panel, anchor=north west
                ] (a) at (0,  3.9) {
				    \input{mnist_parts.pgf}
                };
                \node[label] at (a.north west) {A};
                \node[panel, anchor=north west
                ] (b) at (5.2,  3.7) {
                    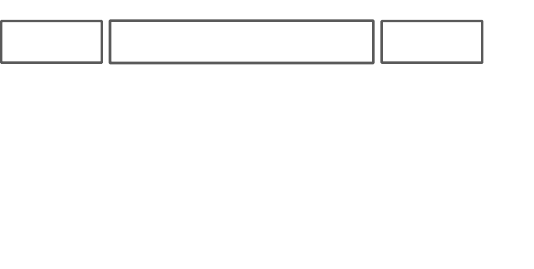
                };
                \node[label] at (5,  3.9) {B};
                \node[panel, anchor=north west
                ] (c) at (14.5,  3.9) {
                    \input{mnist_obsv.pgf}
                };
                \node[label] at (14.3,  3.9) {C};
        \end{tikzpicture}
        \caption{%
		\textbf{(A)} 
		Schematic network topology for a network of 28\texttimes{}28 $\rightarrow$ 256 $\rightarrow$ 10 neurons.
		Partitions that can be run consecutively on hardware are marked.
		The four partitions in the first layer are interchangeable.
        \textbf{(B)}
		Data flow of the model from (A) using five partitions, where the additional need to record and play back events to/from the host computer in-between layers is visualized by dashed lines.
        \textbf{(C)} Measured spikes and membrane potentials of each hardware run.
        To run the fifth partition, the spikes from the first four partitions need to be known.
        On a multi-chip setup with at least five chips, all parts could be run in parallel.
		}
        \label{fig:mnist}
\end{figure}

Executing the network described in \cref{sec:mnist_methods} with the single-chip \gls{bss2} system is only possible after partitioning it into five parts as the $28\times 28$ inputs require multiple neuron circuits to be connected, see \cref{fig:mnist}A.
Specifically, the \num{784} pixels are mapped to the same number of signed weights per neuron, requiring two hardware synapses each, thereby requiring eight\footnote{actually seven, but to simplify the mapping onto \gls{bss2}, eight circuits are used.} combined neuron circuits.
By partitioning the hidden layer of 256 units into four parts, the \num{64} units per partition comply with the \gls{bss2} substrate ($64 \times 8 = 512$, the number of neuron circuits on the chip) so that each of the parts can be executed in one run.
For each run, the input events need to be provided, as indicated by the dashed lines in \cref{fig:mnist}B, which showcases a schematic view of the network and the necessary partitions for execution on \gls{bss2}.
Once the spike events have been read out from the four parts of the hidden layer they are reassembled in software which is required to emulate the readout layer.
The observed spikes on \gls{bss2} for each partition and the membrane traces of the output layer are shown in \cref{fig:mnist}C.

The particular \gls{ttfs} encoding used here assigns spike times $t_i^s$ to pixel values $x_i$ in a linear manner,
\begin{align}
    x_i \rightarrow t_i^s = \left(T - \left\lfloor \frac{T}{\delta t} \cdot\frac{x_i - x_\text{min}}{x_\text{max}-x_\text{min}}\right\rceil \cdot \delta t\right)
\end{align}
where $T$ is the sequence length per image, that together with the time interval $\delta t$ determines the encoding resolution.
The mixed flooring and ceiling brackets indicate rounding to the next integer and $x_\text{min/max}$ are the minimum/maximum pixel values of the dataset.
All previous publications reporting on this benchmark on \gls{bss2} used a scaled-down image size of $16 \times 16$ to reduce input dimensionality in order to fit the whole network on a single chip instance, compare \cref{tab:mnist_comparison}.
Our model is the first implementation using the full resolution of $28 \times 28$ on \gls{bss2} ---and a slightly larger hidden layer (\num{256} compared to \num{246} before; see \cref{fig:mnist}A)---, and reaches \SI{97.9(1)}{\percent} using similar training methods.
Although the slight improvement in classification performance does not indicate the necessity for the development of means to run larger-scale models, it represents an important milestone in the validation of our implementation and hardware operation against previous results.

\begin{table}
    \centering
    \small
    \caption{MNIST Experiment}
    \label{tab:mnist_comparison}
    \begin{tabular}[t]{l|cc}
        \toprule
        \textbf{Publication} & \textbf{Input Size} & \textbf{Test Accuracy [\%]} \\
        \midrule
        \citet{goeltz2021fast} & $16 \times 16$ & $96.9 \pm 0.1$ \\
        \citet{cramer2022surrogate} & $16 \times 16$ & $97.6 \pm 0.1$ \\
        \textbf{This work} & $28 \times 28$ & $\mathbf{97.9 \pm 0.1}$ \\
        \bottomrule
    \end{tabular}
\end{table}

\subsubsection{EuroSAT}\label{subsec:eurosat}

\begin{figure}
    \centering
    \tikzset{
             panel/.style={
                     inner sep=0pt, outer sep=0, execute at begin node={\tikzset{anchor=center, inner sep=.33333em}}
             },
                label/.style={anchor=north west, inner sep=0, outer sep=0}
        }
        \begin{tikzpicture}
                \node[panel, anchor=north west
                ] (a) at (0,  4.0) {
                    \input{encodings.pgf}
                };
                \node[label] at (0, 4.1) {A};
                \node[panel, anchor=north west
                ] (b) at (8.7, 4.65) {
                    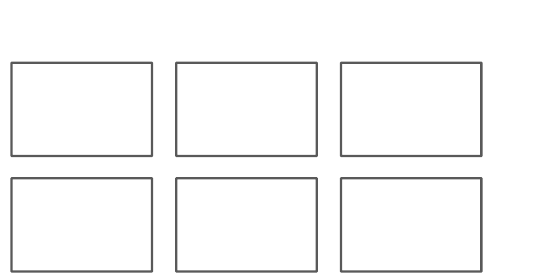
                };
                \node[label] at (8.5,  4.1) {B};
        \end{tikzpicture}
        \caption{\textbf{(A)} 
        (left) Example image of the EuroSAT dataset.
        (middle) The image \gls{ttfs} encoded.
        \textbf{(B)}
        Partitioning and placement of the network used to classify the EuroSAT dataset.
		The basic synapse and neuron layout of the \gls{bss2} \acrshort{asic} is shown in each column:
		in the center, two rows of neuron circuits are located;
		each neuron row is fed from the adjacent synapse array (top/bottom rectangles).
		Neuron circuits can be combined to form larger logical neurons, supporting larger fan in.
        On the left of each hardware instance, the source and size of the fan-in are indicated.
        Each neuron in the first hidden layer has a receptive field of $3 \times 3 \times 3$ and can be mapped to one \gls{bss2} instance.
		To reduce the number of input spikes, we run it in multiple parts (indicated by the red box).
        The neurons in the second layer consist of four connected neuron circuits.
        This layer, as well as the readout layer, is executed in a single run each.}
        \label{fig:eurosat}
\end{figure}

We trained the model described in \cref{sec:methods:eurosat} to classify the EuroSAT dataset~\citep{helber2017eurosat}.
Its partitioning and placement on \gls{bss2} is visualized in \cref{fig:eurosat}B.
Instead of densely projecting the large input space onto the first hidden layer, each neuron in the layer has a small receptive field of $3 \times 3 \times 3$ pixels.
The receptive fields are moved over the spatial coordinates (height and width) of the image with stride \num{3}, resulting in each neuron receiving a unique set of input pixels.
For the \gls{bss2} system this encoding is particularly convenient since it makes uses of the system's intrinsic support for placing sparse connections.
With the given size of the receptive field, the first hidden layer has a size of \num{484} neurons with \num{27} inputs each.
Each synapse row on \gls{bss2} can distinguish \num{64} event labels, hence, we uniquely address a maximum of 64 neurons through the same row.
This allows to map the sparse projection in blocks of $27 \times 64$ ``signed'' hardware synapses onto \gls{bss2} and thus run the whole first layer at once.
The large input space in conjunction with the used \gls{ttfs} encoding scheme still results in a fair amount of spikes, hence, means for reducing the number of input events are applied --- also by partitioning of the first hidden layer, thereby reducing the number of input neurons required per execution (see red box in \cref{fig:eurosat}B).
We execute this layer in \num{8} parts, resulting in \num{10} runs needed to emulate the whole network.
The remaining projections between layers have all-to-all connectivity.
The second hidden layer of size \num{128}, can be emulated within one run by connecting four neuron circuits on \gls{bss2} to form one neuron in order to support a fan-in of \num{484} from the previous layer.
The readout layer is implemented with single-circuit neurons.

To avoid the on-chip spike event rate to exceed the system's bandwidth, we use an \gls{ttfs} input encoding scheme, see \cref{fig:eurosat}A.
Each pixel value $x_i \in [0, 1]$ is interpreted as a constant current onto a \gls{lif} neuron with an infinite refractory period, i.e.\ the neuron can only spike once at $t_i^\text{s}$ (cf.\ \cite{cramer2022surrogate}).
This yields an early spike time for stronger pixel intensities and no input spike if the pixel value is too small.
We add a bias value $x_\text{min}$ to $x_i$ to bias the inputs towards early spiking.
The spike times $t_i^\text{s}$ are numerically computed according to
\begin{align}\label{eq:encoding_eurosat}
	x_i \rightarrow t_i^\text{s} = t\vert_{v_i(t) = \vartheta_\text{en}} \quad \text{with} \quad \dot{v_i}(t) = -\frac{1}{\tau_\text{en}}v_i(t) + x_i + x_\text{min},
\end{align}
with $v_i$ being a membrane state, and $\vartheta_\text{en}$ a threshold.
See \cref{fig:eurosat}A for an example.
Using this encoding, we achieve an average spike count per time bin of \num{162} (averaged over training set and time bins) and the maximum average spike count encountered in a bin (averaged over training set) to \num{527}.

The \gls{bss2} \gls{fpga} only processes two spikes per clock cycle, i.e.\ simultaneous sends might get delayed.
If the maximum bandwidth is exceeded for longer time spans, spikes are dropped.
To minimize simultaneous events, we compute the spike times at FPGA resolution.
However, since the dataset is constituted of only 252 unique pixel values only the same number of unique spike times will occur.
In the forward direction, we therefore jitter the pixel images by adding Gaussian noise, $x_i + \mathcal{N}(\mu = 0, \sigma_\text{in})$.
For gradient optimization we assume the same resolution $\delta t$ as in the simulations.
All parameters are summarized in \cref{tab:eurosat_params}.

\begin{figure}
    \centering
    \begin{minipage}[c]{0.50\textwidth}
	    \input{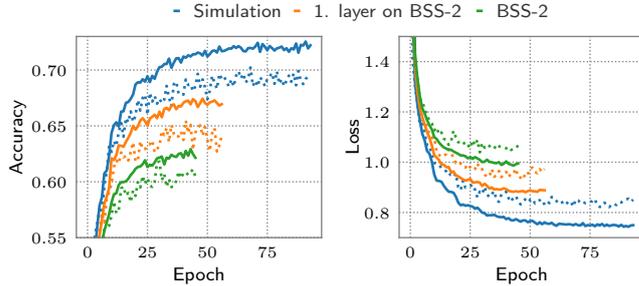}
    \end{minipage}\hfill
    \begin{minipage}[c]{0.45\textwidth}
        \vspace{-10px}
        \caption{%
            Accuracy (left) and loss (right) of the model on the EuroSAT dataset in simulation and/or on \gls{bss2}.
            The solid lines correspond to the training set, the dotted to the validation set.
            Blue corresponds to a fully simulated network, green to the whole \gls{snn} partitioned emulated on \gls{bss2}, and orange to mixed simulation/\gls{bss2} execution with only the first layer being emulated on \gls{bss2}.
        } \label{fig:ttfs:performance}
    \end{minipage}
\end{figure}

In \cref{fig:ttfs:performance} we show the training (solid) and validation (dotted) accuracy and loss of our model on the EuroSAT dataset.
We achieve a test accuracy of \SI{69.5}{\percent} (blue) in a software-only training.
When emulating the whole model on \gls{bss2} (green) the test accuracy is \SI{61.9}{\percent}.
We showcase an example of mixed numerical simulation \slash{} \gls{bss2} emulation where only the first hidden layer is run on \gls{bss2} (orange).
A penalty of approximately \SI{7.5}{\percent} is observed on \gls{bss2}, with approximately \SI{50}{\percent} of this value attributable to the first hidden layer, as indicated by the mixed simulation/BSS-2 experiment. 
This emphasizes the importance of support for mixed execution to investigate and improve the performance of future models and systems.

\section{Discussion}\label{sec:discussion}

This paper emphasizes the role of software in enabling the partitioned emulation of large-scale \glspl{snn} on the \gls{bss2} neuromorphic substrate.
While manual partitioning of suitable \gls{snn} topologies has always been a viable approach, the integration of software support into the \gls{bss2} software stack enables researchers to shift their focus from system handling to modeling.
The present work is concerned with enabling the expression of manually partitioned networks, with the aim of enabling a rapid adoption by modelers.
Future developments will aim to provide automated algorithms for partitioning, thereby relieving users of this task and enabling the creation of more complex partitioned network topologies.

We demonstrated partitioned emulation on \gls{snn} models classifying the MNIST and EuroSAT datasets, which require the use of many single \gls{bss2} chip instances.
While the training processes used surrogate gradient-based learning methods \citep{neftci2019surrogate}, an event-driven training approach, e.g., using the EventProp algorithm \citep{wunderlich2021event} and the event-driven \gls{bss2} modeling \gls{api} \texttt{jaxsnn} \citep{mueller2024jaxsnn}, could provide further efficiency gains by exploiting sparsity in observables, thereby minimizing data transfers between host and neuromorphic hardware, as well as in numerical computations.

To validate our implementation, we used the MNIST dataset, as there are several publications using single-chip \gls{bss2} systems.
Our model performs slightly better on $28 \times 28$ image resolution than the smaller models on $16 \times 16$ images, achieving \SI{97.9(1)}{\percent} test accuracy.
For further details, please see \cref{sec:results_mnist}.
This represents the best performance on MNIST recorded on \gls{bss2} to date.
We acknowledge that this improvement may also be partially attributable to a more efficient input encoding and training setup.
This is the first time the full-scale benchmark has been run on \gls{bss2}.
The capacity to benchmark systems without the necessity for extensive pre-processing and downscaling ensures fair comparison to other systems, thereby underscoring the importance of facilitated partitioned emulation of \glspl{snn} on small-scale systems.

For the larger EuroSAT task, we present the first results obtained on \gls{bss2}.
We showcase the emulation of the largest \gls{snn} to date on \gls{bss2} through the partitioning into subnetworks, each of which is executable on the available hardware substrate.
The sparse input projection enables us to map a 12288-dimensional input space to the hardware.
Due to connectivity sparsity, the first hidden layer is emulated in eight parts, resulting in ten partitions for the whole network.
In the future, sufficiently large multi-chip systems will be capable of emulating all partitions concurrently.
The sequential execution of the model on \gls{bss2} resulted in a test accuracy of \SI{61.9}{\percent}, thus supporting our presented approach for large-scale model emulation.
The performance gap to the numeric simulation is assumed to be not intrinsic to the analog nature of the system.
Potential causes for the observed performance degradation on \gls{bss2} include suboptimal hardware operation points and training setup, in addition to spike loss in the input layer due to bandwidth constraints.
We are optimistic to resolve the latter by stretching the experiment in time to minimize the number of simultaneous events and by increasing the number of partitions of the first hidden layer.
Our support for emulating only parts of the network on \gls{bss2} and numerically simulating the remaining parts is a crucial feature for identifying hardware-specific intricacies and debugging the model's performance, e.g., by identifying which dynamics of the \gls{snn} are emulated at a suboptimal hardware operation point.

While partitioned emulation is typically superlinearly slower than on a sufficiently large substrate, the ability to explore larger networks is valuable, especially when considering typical hardware development cycle times and costs.
We have shown this superlinearity for the MNIST experiment, where the inter-execution data transfer via the host however is insignificant, leaving the linear scaling to the preparation, execution and post-processing of the sequential executions.

Due to the mixed-signal nature of the \gls{bss2} architecture ---and many other neuromorphic systems \citep{thakur2018mimicthebrain_nourl}--- the partitioning of \glspl{snn} does not affect the emulation fidelity compared to a system with network-matching system size:
spikes are events in time that can be reliably recorded (within the constraints of the system's I/O bandwidth) and played back at later points in time, thereby providing deterministic communication between subnetworks.
The ability to facilitate answering questions about the desired model and hardware system size with the confidence of a realistic emulation is a key outcome of this work.
This not only addresses the immediate need to understand the behavior of larger networks on existing hardware, but also provides valuable insight into the feasibility and performance expectations for future, more expansive ---and expensive--- neuromorphic systems.

\section*{Author Contributions}\label{sec:author_contributions}

EA~\&~JVS: Investigation, visualization, methodology, software, writing --- original draft, writing --- reviewing \& editing;
EA: Conceptualization;
PS~\&~EM: Conceptualization, methodology, software, writing --- original draft, writing --- reviewing \& editing;
EM: Supervision;
DD~\&~GM: Methodology, resources, validation, writing --- original draft \& editing.
JS: Methodology, supervision, resources, funding acquisition, writing — reviewing \& editing.

\label{sec:acknowledgements}
The authors wish to thank all present and former members of the Electronic Vision(s) research group contributing to the \acrlong{bss2} neuromorphic platform.

\noindent%
\subsection*{Funding}
This work has received funding from
the EC Horizon 2020 Framework Programme
under grant agreements
785907 (HBP SGA2) 
and
945539 (HBP SGA3), 
the EC Horizon Europe Framework Programme
under grant agreement
101147319 (EBRAINS 2.0),
the \foreignlanguage{ngerman}{Deutsche Forschungsgemeinschaft} (DFG, German Research Foundation) under Germany’s Excellence Strategy EXC 2181/1-390900948 (the Heidelberg STRUCTURES Excellence Cluster),
the German Federal Ministry of Education and Research under grant number 16ES1127 as part of the \foreignlanguage{ngerman}{\emph{Pilotinnovationswettbewerb `Energieeffizientes KI-System'}},
the Helmholtz Association Initiative and Networking Fund [Advanced Computing Architectures (ACA)] under Project SO-092,
and the \foreignlanguage{ngerman}{Lautenschläger-Forschungspreis} 2018 for Karlheinz Meier.
This study has been supported by the European Space Agency's Ariadna scheme (Study Ref.\ 4000136024/21/NL/GLC/my).

\section*{Conflict of Interest Statement}

The authors declare that the research was conducted in the absence of any commercial or financial relationships that could be construed as a potential conflict of interest.

\section*{Supplemental Data}
\Cref{tab:mnist_params} and \cref{tab:eurosat_params} provide parameters for the MNIST (\cref{sec:results_mnist}) and EuroSAT (\cref{subsec:eurosat}) experiments.

\begin{table}
    {
    \centering
    \parbox[t]{.49\linewidth}{
        \small
        \caption{MNIST Experiment}
        \label{tab:mnist_params}
        \begin{tabular}[t]{lc}
            \toprule
            \textbf{Training} & \\
            Batch size & 100 \\
            Learning rate & 0.002 \\
            Epochs & 100 \\
            Learning rate decay & 0.985 \\
            Vertical/Horizontal flip & \SI{25}{\percent} probability\ \\
            Dropout & 0.15 \\
            Optimizer & Adam (default) \\
            SuperSpike slope $\alpha$ & 50 \\
            \midrule
            \textbf{Simulation \& Gradient} & \\
            $\delta t$ & \SI{1}{\micro\second} \\
            Time steps $T$ & 30\\
            Leakage potential & \num{0} \\
            Reset potential & \num{0} \\
            Threshold & \num{1} \\
            Mem. time constant & \SI{6}{\micro\second} \\
            Syn. time constant & \SI{5.7}{\micro\second} \\
            Readout scale & 3 \\
            \midrule
            \textbf{Regularization} & \\
            Bursts & 0.0025 \\
            $\Theta_\text{h}$ & 0.0033 \\
            $\Theta_\text{o}$ & 0.0033 \\
            $v_\text{o}$ & 0.00016 \\
            $\gamma$ & 0.985 \\
            \midrule
            \textbf{Encoder} & \\
            $x_\text{min}$ & 0 \\
            $x_\text{max}$ & 1 \\
            \midrule
            \textbf{BSS-2 Operation Point}$^\Delta$ & \\
            \texttt{i\_synin\_gm}$^*$ & [\num{800}, \num{400}] \\
            \texttt{synapse\_dac\_bias}$^*$ & [\num{850}, \num{700}] \\
            \texttt{leak} & \num{80} \\
            \texttt{reset} & \num{80} \\
            \texttt{threshold} & \num{120} \\
            \texttt{membrane\_capacitance} & \num{63} \\
            \texttt{refractory\_time} & \SI{1}{\micro\second} \\
            \bottomrule
        \end{tabular}
    }
    \hfill
    \parbox[t]{.49\linewidth}{
        \small
        \caption{EuroSAT Experiment}
        \label{tab:eurosat_params}
        \begin{tabular}[t]{lc}
            \toprule
            \textbf{Training} & \\
            Batch size & 64 \\
            Learning rate & 0.001 \\
            Max. epochs & 25 / 15 \\
            Learning rate decay & 0.5 every $10, \dots, 60$ \\
            Vertical/Horizontal flip & \SI{50}{\percent} probability\ \\
            Optimizer & Adam (default) \\
            \midrule
            \textbf{Simulation \& Gradient} & \\
            $\delta t$ & \SI{1}{\micro\second} \\
            Time steps $T$ & 64 \\
            Leakage potential & 0 \\
            Reset potential & 0 \\
            Syn. time const.$^*$ & $[10, 10, 10]$\si{\micro\second} \\
            Mem. time const.$^*$ & $[10, 10, 10]$\si{\micro\second} \\
            Thresholds $\vartheta$$^*$ & $[1, 1, -]$ \\
            SuperSpike slope $\alpha$$^*$ & $[10, 10, -]$ \\
            \midrule
            \textbf{Encoder} & \\
            Threshold $\vartheta_\text{en}$ & 0.32 \\
            Time constant $\tau_\text{en}$ & $\SI{20}{\micro\second}$ \\
            $x_\text{min}$ & 0.1 \\
            $\sigma_\text{in}$ & 0.003 \\
            \midrule
            \textbf{BSS-2 Operation Point}$^\Delta$ & \\
            \texttt{i\_synin\_gm}$^*$ & [\num{350}, \num{350}, \num{300}] \\
            \texttt{synapse\_dac\_bias}$^*$ & [\num{1000}, \num{1000}, \num{600}] \\
            \texttt{leak}$^*$ & [\num{100}, \num{100}, \num{120}] \\
            \texttt{reset}$^*$ & [\num{100}, \num{100}, \num{120}] \\
            \texttt{threshold}$^*$ & [\num{160}, \num{160}, \num{140}] \\
            \texttt{membrane\_capacitance} & \num{63} \\
            \texttt{refractory\_time}$^*$ & [\num{1}, \num{1}, \num{0.4}]\si{\micro\second} \\
            \bottomrule
        \end{tabular}
    }
	}
	\vspace{.25\baselineskip}
	\\
	\footnotesize{%
		$^\Delta$ Integer numbers are either
		digitally settable to a set of specific values (e.g., \texttt{membrane\_capacitance}),
		digitally settable to a range of values (e.g., \texttt{i\_synin\_gm}),
		or calibrated and given in on-chip measured ``ADC'' units (e.g., \texttt{threshold}) as used by the calibration library \texttt{calix}.\\
		$^*$ List index corresponds to layer index.
	}\\
\end{table}

\section*{Data Availability Statement}
Publicly available datasets were analyzed in this study.
The EuroSAT dataset can be found here: \url{https://github.com/phelber/EuroSAT}.
Researchers can use the EBRAINS research infrastructure to access BrainScaleS-2 systems: \url{https://www.ebrains.eu/nmc}.
An MNIST example can be found in the BrainScaleS-2 tutorial collection: \url{https://electronicvisions.github.io/documentation-brainscales2/latest/brainscales2-demos}.

\printbibliography

\end{document}